\documentclass[11pt]{article}

\usepackage[preprint]{acl}

\usepackage{times}
\usepackage{latexsym}

\usepackage[T1]{fontenc}

\usepackage[utf8]{inputenc}

\usepackage{microtype}

\usepackage{inconsolata}

\usepackage{graphicx}
\usepackage{microtype}
\usepackage{hyperref}
\usepackage{url}
\usepackage{booktabs}

\usepackage{lineno}

\usepackage{array}
\usepackage{tabularx}
\usepackage{graphicx}
\usepackage{tcolorbox}
\usepackage{tikz}
\usepackage{amsmath}
\usepackage{graphicx}
\usepackage{amssymb}
\usepackage{multirow}
\usepackage{wrapfig,lipsum,booktabs}
\usepackage{tabu}
\usepackage{graphicx}
\usepackage{subfigure}
\usepackage{tabu}
\usepackage{enumitem}
\usepackage{makecell}
\usepackage{mathtools}
\usepackage{amsthm}
\usepackage{hyperref}
\usepackage[table]{xcolor}
\usepackage{pgfplots}
\usepackage{pgfplotstable}
\usepackage[dvipsnames]{xcolor}
\usepackage{inconsolata}
\usepackage{pgfplots, stackengine}
\pgfplotsset{compat=1.18} 
\usepgfplotslibrary{groupplots} 
\usetikzlibrary{positioning}

\usepackage[linesnumbered,ruled,vlined]{algorithm2e}
\usepackage{amsmath, amssymb, amsthm}
\usepackage{mathtools}

\usepackage{algorithmic}
\title{\includegraphics[height=1.0em]{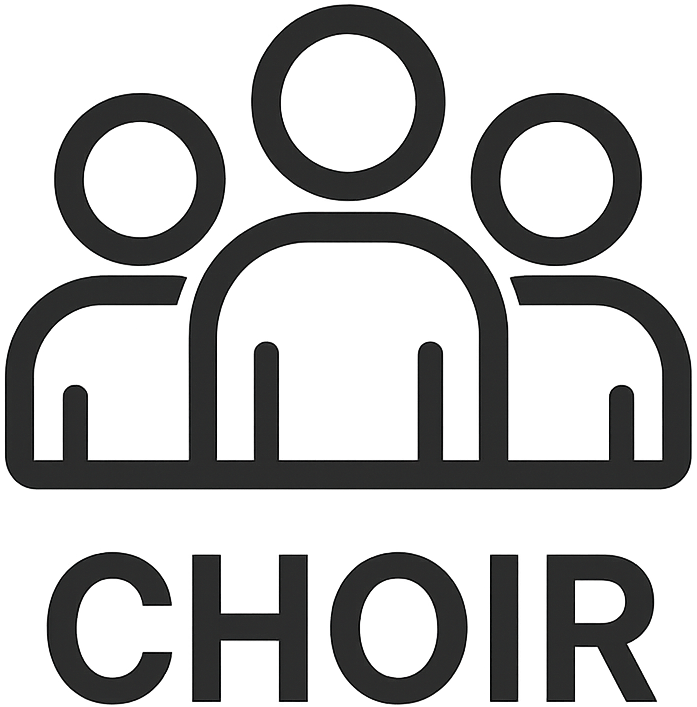} CHOIR: Collaborative Harmonization fOr Inference Robustness}

\author{Xiangjue Dong\quad Cong Wang\quad Maria Teleki\quad Millennium Bismay\quad James Caverlee \\
  Texas A\&M University \\
  \texttt{\{xj.dong, congwang, mariateleki, mbismay, caverlee\}@tamu.edu}}

\begin{document}
\maketitle
\begin{abstract}
Persona-assigned Large Language Models (LLMs) can adopt diverse roles, enabling personalized and context-aware reasoning. However, even minor demographic perturbations in personas, such as simple pronoun changes, can alter reasoning trajectories, leading to divergent sets of correct answers. Instead of treating these variations as biases to be mitigated, we explore their potential as a constructive resource to improve reasoning robustness.
We propose CHOIR (Collaborative Harmonization fOr Inference Robustness), a test-time framework that harmonizes multiple persona-conditioned reasoning signals into a unified prediction. CHOIR orchestrates a collaborative decoding process among counterfactual personas, dynamically balancing agreement and divergence in their reasoning paths.
Experiments on various reasoning benchmarks demonstrate that CHOIR consistently enhances performance across demographics, model architectures, scales, and tasks -- without additional training. Improvements reach up to 26.4\% for individual demographic groups and 19.2\% on average across five demographics. It remains effective even when base personas are suboptimal. By reframing persona variation as a constructive signal, CHOIR provides a scalable and generalizable approach to more reliable LLM reasoning.
\end{abstract}

\begin{figure}[t]
    \centering
    \includegraphics[width=0.99\columnwidth]{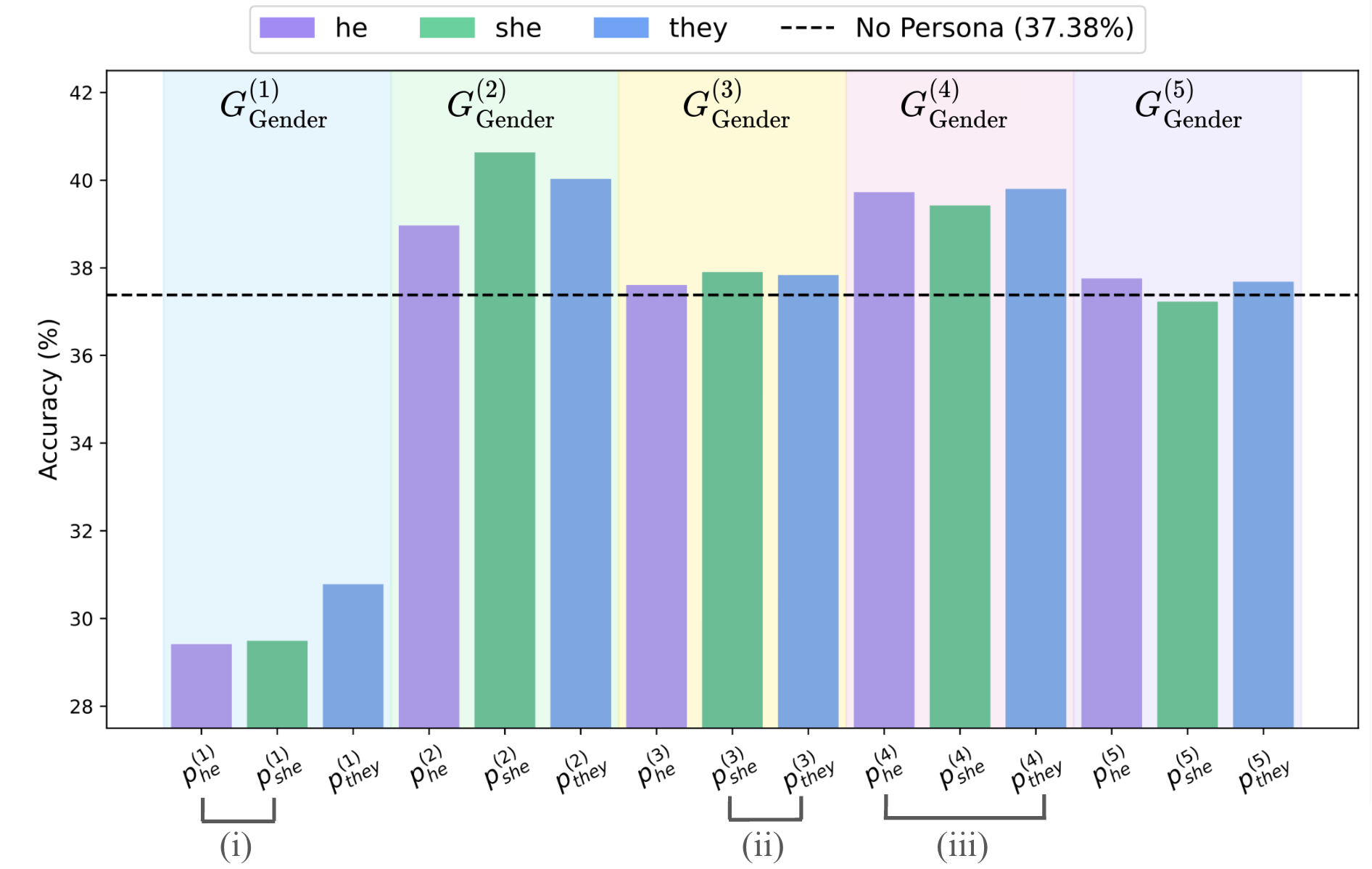}
    \caption{\textbf{LLM performance is sensitive to simple pronoun perturbations.} This figure shows accuracy variations of the Llama-1B model on GSM8K across five different base personas, each perturbed with \textit{he}, \textit{she}, and \textit{they} pronouns. Even when overall accuracies are similar (pairs i, ii, iii), the sets of correctly solved problems differ (Figure \ref{fig:two}). Personas for $G_{\text{Gender}}^{(i)}$ are listed in Table~\ref{tab:gender_group}; detailed analysis is in \S\ref{ssec:perturbation}.}
    \label{fig:one}
    \vspace{-15pt}
\end{figure}


\section{Introduction}

Persona-assigned Large Language Models (LLMs) can improve zero-shot reasoning by steering the model to generate personalized, in-character responses with detailed reasoning traces~\citep{role-play-better-zero-shot,two-tales, personamath, roleplay-paradox, lego}. However, persona assignment also introduces challenges: it can amplify biases, produce unsafe outputs \citep{roleplay-paradox, personas-not-helpful}, or expose deep-rooted biases inherent in LLMs \citep{gupta2024bias}. These risks highlight the need for careful implementation and improved methods to detect and mitigate biases, particularly given the limited success of existing techniques \citep{gupta2024bias}.

\begin{figure}[t]
    \centering  
    \includegraphics[width=0.99\columnwidth]{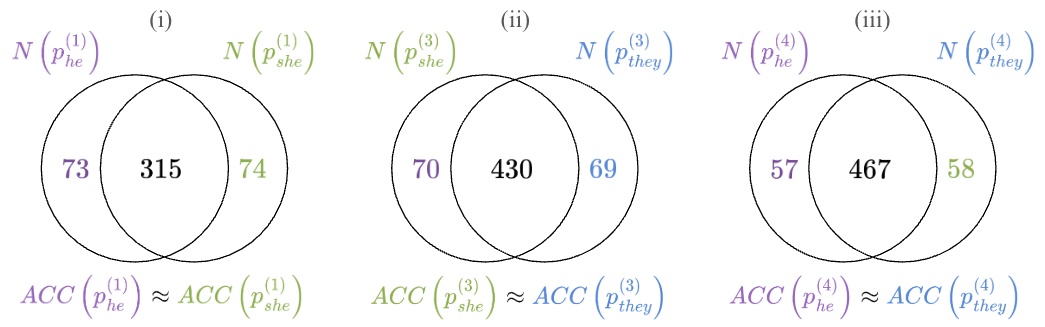}
    \caption{\textbf{Similar overall accuracy reveals complementary reasoning strengths.} This figure shows the overlap in correct answers for three pairs of personas from Figure \ref{fig:one} with nearly identical accuracy. For example, in (i), the \textit{he} and \textit{she} personas differ by only 0.1\% in accuracy yet share 315 correct answers, with 73 and 74 uniquely solved, respectively. These non-overlapping sets demonstrate that minimal perturbations can lead to distinct reasoning paths (See \S\ref{ssec:accuracy} for details.)}
    \label{fig:two}
    \vspace{-15pt}
\end{figure}

To better understand model sensitivity to demographic cues embedded within persona prompts, we construct a set of counterfactual gender personas by perturbing demographic terms. Our analysis reveals two key findings: (1) even simple demographic perturbations can significantly affect model performance (Figure~\ref{fig:one}), echoing recent findings on persona-induced bias~\cite{gupta2024bias}; and (2) while overall accuracy remains similar, different persona framings lead to divergent sets of correct answers (Figure~\ref{fig:two}). These findings indicate that even small perturbations in personas can lead to different reasoning trajectories.

While much of the existing work treats such variation as a form of bias to be diagnosed and mitigated, we take a different perspective: \textbf{Can demographic perturbations be used constructively to support more robust reasoning?} 
To explore this question, we introduce CHOIR (Collaborative Harmonization fOr Inference Robustness), a test-time framework that enhances reasoning by harmonizing multiple persona-conditioned reasoning paths obtained from diverse counterfactual demographic framing. CHOIR's dynamic weighting mechanism integrates these complementary signals by giving more influence to paths that show strong agreement with the rest of the ensemble, allowing the model to leverage diverse perspectives without overcommitting to any single one.

CHOIR operates entirely at inference time and requires no additional training. Experiments across diverse reasoning datasets, model architectures, and scales show that it consistently enhances both accuracy and robustness over baselines, achieving up to 26.4\% improvement for individual demographic groups and an average of 19.2\% across five demographics (\textit{Gender}, \textit{Race}, \textit{Religion}, \textit{Disability}, and \textit{Age}). Importantly, CHOIR remains effective even when base personas are noisy or suboptimal and generalizes well across demographics.

These results indicate that persona variation, when handled systematically, can serve as a rich resource rather than a liability. CHOIR provides a simple and general framework for decoding-time collaboration among demographic framings, opening up new directions for socially aware, reliable, and robust LLM reasoning.

\begin{table}[t]
\begin{center}
\resizebox{0.99\columnwidth}{!}{%
\begin{tabular}{p{1cm}p{11.2cm}}
\toprule
\multicolumn{1}{c}{\bf $G_{\text{Gender}}^{(i)}$}  &\bf Personas ($p^{(i)}_{\text{he}}$, $p^{(i)}_{\text{she}}$, $p^{(i)}_{\text{they}}$) \\
\midrule
$G_{\texttt{Gender}}^{(1)}$ & \texttt{a hardworking construction worker tirelessly petitioning for \{his/her/their\} sibling's immigration case} \\
$G_{\texttt{Gender}}^{(2)}$ & \texttt{a retired basketball coach with decades of experience, known for \{his/her/their\} expertise in developing point guards} \\
$G_{\texttt{Gender}}^{(3)}$ & \texttt{a single \{man/woman/person\} who takes \{his/her/their\} personal development and emotional health seriously} \\
$G_{\texttt{Gender}}^{(4)}$ & \texttt{a retired childcare professional who always has a soothing lullaby up \{his/her/their\} sleeve}\\
$G_{\texttt{Gender}}^{(5)}$ & \texttt{a music producer who collaborates with Bette Midler on \{his/her/their\} new album} \\
\bottomrule
\end{tabular}}
\end{center}
\caption{Counterfactual gender persona group $G_{\text{Gender}}^{(i)}$ used in Figure~\ref{fig:one} and \ref{fig:two}.
}\label{tab:gender_group}
\vspace{-15pt}
\end{table}

Our contributions are threefold: 

\begin{itemize}[leftmargin=2em, topsep=0pt,itemsep=0pt, partopsep=0pt] 
\item We show that even minor demographic perturbations in personas can lead to meaningful divergences in reasoning outputs (\S \ref{sec:analysis}).
\item We propose \textbf{CHOIR}, a test-time decoding framework that dynamically harmonizes these diverse reasoning paths to improve both robustness and accuracy (\S \ref{sec:method}).
\item Through extensive experiments, we show that CHOIR consistently outperforms strong baselines and its performance advantage grows as model scale increases (\S \ref{sec:experiments}).

\end{itemize}

\begin{table*}[!t]
\begin{center}
\resizebox{0.99\textwidth}{!}{%
\begin{tabular}{llp{13cm}}
\toprule
\bf $G_{\text{Attribute}}$  & \bf Demographic Term ($w_j$) & \bf Persona ($p_{w_j}$) \\
\midrule
$G_{\text{Gender}}$ & \texttt{his}, \texttt{her}, \texttt{their} & \texttt{a hardworking construction worker tirelessly petitioning for \{\} sibling's immigration case} \\
$G_{\text{Race}}$ & \texttt{White}, \texttt{Black}, \texttt{Asian} & \texttt{a/an \{\} social psychologist who used to work in law enforcement} \\
$G_{\text{Religion}}$ & \texttt{Jewish}, \texttt{Christian}, \texttt{atheist} & \texttt{a/an \{\} living in New York City} \\
$G_{\text{Disability}}$ & \texttt{disabled}, \texttt{able-bodied} & \texttt{a/an \{\} individual who has personally experienced the challenges of navigating the town and offers valuable insights and suggestions} \\
$G_{\text{Age}}$ & \texttt{old}, \texttt{young} & \texttt{an/a \{\} philosophy professor who is skeptical of quantitative measures for qualitative concepts}\\
\bottomrule
\end{tabular}}
\end{center}
\caption{\textbf{Counterfactual demographic persona group $G_{\text{Attribute}}$.} Each group varies by a specific demographic term ($w_j$) and includes corresponding persona templates ($p_{w_j}$) used in our analysis.}
\label{tab:demographics}
\vspace{-10pt}
\end{table*}

\section{Uncovering Complementary Knowledge in Counterfactual Personas}
\label{sec:analysis}
\subsection{Simple Demographic Perturbation Impacts Performance}
\label{ssec:perturbation}

To evaluate the model's sensitivity to minor demographic cues embedded in persona prompts, we construct a set of counterfactual gender personas through the perturbing of demographic terms. Each group, $G_{\text{Gender}}^{(i)}$, contains three parallel personas, $(p^{(i)}_{\text{he}}, p^{(i)}_{\text{she}}, p^{(i)}_{\text{they}})$, that differ only in the pronouns \textit{he}, \textit{she}, and \textit{they}. One base persona (e.g., $p^{(i)}_{\text{he}}$) is randomly sampled from the PersonaHub dataset\footnote{\url{https://huggingface.co/datasets/proj-persona/PersonaHub}}, while the other two are created by systematically perturbing demographic terms (see Table~\ref{tab:gender_group}). This design ensures that all three personas share the same underlying identity and structure, differing only in their expressed demographic identity. We then use these persona groups to ask: \textbf{Are these gender-related expressions minor surface-level variations, or do they meaningfully influence the model's reasoning path?}

Figure \ref{fig:one} presents results for 15 personas, drawn from 5 groups where each group includes one base persona and its two counterfactual demographic variants (Table~\ref{tab:gender_group}), evaluated using the Llama-1B-Instruct model~\citep{touvron2023llama} on the GSM8K dataset~\citep{gsm8k}. Notably, even personas within the same group, differing only by surface-level demographic cues, often exhibit non-trivial performance gaps. In particular, in $G_{\text{Gender}}^{(2)}$, the persona $p^{(2)}_{\text{she}}$ achieves more than $2\%$ higher accuracy than both $p^{(2)}_{\text{he}}$ and $p^{(2)}_{\text{they}}$. \textbf{These findings indicate that even small demographic perturbations can influence reasoning performance.} While these results are reported for Llama, we observe qualitatively similar trends in other models, indicating that persona-induced variance is a general phenomenon across architectures, consistent with prior observations of persona-related bias~\cite{gupta2024bias}.

\subsection{Similar Accuracy, Divergent Reasoning}
\label{ssec:accuracy}

Accuracy alone, however, does not capture the full picture. To probe deeper, we compare the sets of correctly answered questions across three persona pairs with nearly identical overall accuracies (Figure~\ref{fig:one}): 
(i) $p^{(1)}_{\text{he}}$ vs. $p^{(1)}_{\text{she}}$ (29.41\% vs. 29.49\%), 
(ii) $p^{(3)}_{\text{she}}$ vs. $p^{(3)}_{\text{they}}$ (37.91\% vs. 37.83\%), and 
(iii) $p^{(4)}_{\text{he}}$ vs. $p^{(4)}_{\text{they}}$ (39.80\% vs. 39.73\%). 
We ask: \textbf{Do these personas with similar overall accuracy actually solve the same set of questions correctly?}

Figure~\ref{fig:two} illustrates the pairwise overlaps and differences in correct-answer sets for each matched persona pair. 
Surprisingly, even when the persona prompts differ only by a single pronoun, the distribution of correct answers can vary substantially. 
For example, in $G_{\text{Gender}}^{(1)}$, $p^{(1)}_{\text{he}}$ and $p^{(1)}_{\text{she}}$ agree on 315 correct answers, yet each also shows distinct behavior: $p^{(1)}_{\text{he}}$ correctly answers 73 questions that $p^{(1)}_{\text{she}}$ does not, while $p^{(1)}_{\text{she}}$ solves 74 questions that $p^{(1)}_{\text{he}}$ misses. This results in over $10\%$ non-overlapping correct answers, despite only a $0.1\%$ difference in overall accuracy. \textbf{Such discrepancies reveal that even minor demographic perturbations can lead to meaningful divergences in reasoning outputs.}

\smallskip
\noindent \textbf{Takeaways.} Together, these findings indicate that even minimal demographic framing changes can alter the model's behavior in meaningful ways. \textbf{Motivated by this observation, we seek to exploit these variations constructively, allowing different persona demographic variants to complement one another and collectively improve reasoning consistency and robustness.}

\begin{algorithm*}[t]
\caption{CHOIR: Collaborative Harmonization fOr Inference Robustness}
\label{alg:persona-fusion}
\KwIn{Question $x$; base persona $p_1$; persona instruction template $T(\cdot)$; language model $f_\theta$; max generation length $L$; ${\lambda}^{(0)}$}
\KwOut{Answer $y$}

\texttt{// Step 1: Construct Counterfactual Personas}

Construct counterfactual personas $\{p_2, \dots, p_n\}$ by modifying demographic attributes of $p_1$\;

Generate persona-specific instructions: $r_i = T(p_i), \forall i \in \{1, \dots, n\}$\;

\texttt{// Step 2: Dynamic Collaborative Decoding}

\For{each generation step $t = 1$ to $L$}{
    \For{each persona $r_i$ (including base and counterfactual personas)}{
        Compute logits and probabilities: $\mathbf{z}^{(i)}_t = f_\theta(r_i, x, y_{<t}), \mathbf{p}^{(i)}_t = \text{softmax}(\mathbf{z}^{(i)}_t)$\;
        Compute confidence score: $s^{(i)}_t = \max \mathbf{p}^{(i)}_t$\;
    }
    Compute average consensus confidence: $\bar{s}_t = \frac{1}{n} \sum_{i=1}^{n} s^{(i)}_t$\;
    
    \For{each persona $r_i, i \ge 1$}{
        $d^{(i)}_t = |s^{(i)}_t - \bar{s}_t|$ \;
        $\alpha^{(i)}_t = 1 - d^{(i)}_t$ \;
    }
    Set $\alpha^{(0)}_t = \lambda^{(0)}$\;

    Compute base logits from pre-trained knowledge: $\mathbf{z}^{(0)}_t = f_\theta(x, y_{<t})$\;

    Aggregate logits: $\mathbf{z}_{\text{adj},t} = \sum_{i=0}^{n} \alpha^{(i)}_t \cdot \mathbf{z}^{(i)}_t = \alpha^{(0)}_t \cdot \mathbf{z}^{(0)}_t + \sum_{i=1}^{n} \alpha^{(i)}_t \cdot \mathbf{z}^{(i)}_t$\;

    Sample next token: $y_t \sim \mathbf{p}_{\text{final},t} = \text{softmax}(\mathbf{z}_{\text{adj},t})$\;

    \If{$y_t$ is end-of-sequence token}{
        \textbf{break};
    }
}
\Return{$y = \text{post\_processing}\{y_1, \dots, y_t\}$}
\end{algorithm*}

\section{CHOIR: Collaborative Harmonization fOr Inference Robustness}
\label{sec:method}

In this section, we introduce CHOIR (Collaborative Harmonization fOr Inference Robustness), a framework that strengthens the robustness and reliability of persona-assigned LLMs in reasoning tasks. By systematically constructing counterfactual personas and harmonizing their outputs during inference, CHOIR leverages demographic variation as a source of complementary reasoning rather than noise. The method consists of two key stages: Counterfactual Persona Construction and Dynamic Collaborative Decoding.

\subsection{Counterfactual Persona Construction}
The first stage introduces diversity into reasoning by generating counterfactual personas.
Starting from a base persona $p_1$ randomly sampled from the PersonaHub dataset, CHOIR constructs alternative counterfactual personas $\{p_2, \dots, p_n\}$ by modifying the demographic attributes of $p_1$, such as gender pronouns or religious identifiers (Table~\ref{tab:demographics}). These perturbations preserve the underlying identity and task-relevant knowledge while altering surface-level demographic framing.
Each persona $p_i$ is then embedded into an instruction template $T(\cdot)$ (Table~\ref{tab:instruction}), producing persona-conditioned prompts: $r_i = T(p_i), \forall i \in \{1, \dots, n\}$. This process yields a set of parallel reasoning prompts, each representing a distinct demographic lens.



\subsection{Dynamic Collaborative Decoding}

Once the persona instructions $\{r_1, \dots, r_n\}$ are established, CHOIR performs multi-step collaborative inference by harmonizing token-level predictions across personas. The key idea is to first aggregate reasoning signals from all personas and then measure how each individual persona diverges from this collective consensus to determine its stability and contribution. 

At each generation step $t$, the LLM $f_\theta$ processes each persona instruction $r_i$, the input question $x$, the Zero-shot Chain-of-Thought trigger ``Let's think step by step''~\citep{zero_shot_cot}, and the previously generated tokens $y_{<t}$, producing persona-conditioned logits and probabilities: 
\begin{equation}
\small
\mathbf{z}^{(i)}_t = f_\theta(r_i, x, y_{<t}), \quad
\mathbf{p}^{(i)}_t = \text{softmax}(\mathbf{z}^{(i)}_t).
\end{equation}
Following previous work~\cite{geng-etal-2024-survey,kumar-etal-2024-confidence,liu-etal-survey}, we define the confidence of persona $i$ as the maximum token probability: $s^{(i)}_t = \max \mathbf{p}^{(i)}_t$.

\smallskip
\noindent \textbf{Consensus confidence and divergence.} To measure agreement and disagreement across personas, we first compute the average confidence across all personas: $\bar{s}_t = \frac{1}{n} \sum_{i=1}^{n} s^{(i)}_t$, and measure each persona's divergence from this consensus: $d^{(i)}_t = \big| s^{(i)}_t - \bar{s}_t \big|$. 
These differences serve as a proxy for uncertainty in the model's behavior under different demographic perspectives. 

\smallskip
\noindent \textbf{Dynamic weighting and aggregation.}
Next, we assign each persona a dynamic weight based on its consistency with the consensus confidence. The intuition is that small demographic perturbations should not substantially alter the reasoning path: personas that remain close to the collective average are more stable and reliable, whereas those that diverge significantly are less robust. Accordingly, each persona $r_i$ is assigned a weight that decreases with its divergence from the average confidence:
\begin{equation}
\small
\alpha_t^{(i)} =
\begin{cases}
\lambda^{(0)}, &  i = 0 \\
\lambda^{(i)} - d^{(i)}_t, & i \ge 1
.
\end{cases}
\label{eq:weight}
\end{equation}
Here, $\lambda^{(i)}$ is a knowledge weight hyperparameter and $i=0$ corresponds to the non-persona baseline, representing the model's internal pretrained reasoning priors without any demographic influence. The base logits without persona influence are computed as $\mathbf{z}^{(0)}_t = f_\theta(x, y_{<t})$. The specific values chosen for $\lambda^{(i)}$ in our experiments are detailed in \S\ref{ssec:setup}, with a more in-depth analysis of $\lambda^{(0)}$ provided in \S\ref{ssec:rq3}.
The adjusted logits are obtained as a weighted sum of persona-specific and base logits:
\begin{equation}
\small
\begin{aligned}
\mathbf{z}_{\text{adj},t} 
&= \sum_{i=0}^{n} \alpha^{(i)}_t \cdot \mathbf{z}^{(i)}_t \\
&= \alpha^{(0)}_t \cdot 
   \underbrace{\mathbf{z}_t^{(0)}}_{\raisebox{-0.3em}{$\mathclap{\shortstack{\text{\scriptsize Pretrained}\\ \text{\scriptsize Knowledge}}}$}}+ \alpha^{(1)}_t \cdot 
   \underbrace{\mathbf{z}_t^{(1)}}_{\raisebox{-0.3em}{$\mathclap{\shortstack{\text{\scriptsize Base}\\ \text{\scriptsize Persona}}}$}} + \sum_{i=2}^{n} \alpha^{(i)}_t \cdot 
   \underbrace{\mathbf{z}_t^{(i)}}_{\raisebox{-0.3em}{$\mathclap{\shortstack{\text{\scriptsize Counterfactual}\\ \text{\scriptsize Personas}}}$}}.
\end{aligned}
\label{eq:adjusted}
\end{equation}
The final token distribution is computed via softmax: $\mathbf{p}_{\text{final},t} = \text{\text{softmax}}(\mathbf{z}_{\text{adj},t})$, from which the next token $y_t$ is sampled. The decoding process continues iteratively until an end-of-sequence token is generated. Once the complete sequence $y = \{y_1, y_2, \dots, y_t\}$ is generated, CHOIR applies post-processing to extract the final answer~\citep{kong-etal-2024-better}. The full process is summarized in Algorithm~\ref{alg:persona-fusion}.

\section{Experimental Setup}
\label{ssec:setup}

\textbf{Tasks, datasets, and models.}
We evaluate CHOIR on arithmetic reasoning tasks: GSM8K~\citep{gsm8k}, SingleEq~\citep{singleeq}, and MultiArith~\citep{multiarith}, and the commonsense reasoning task: CSQA~\citep{csqa}. Detailed dataset statistics are provided in Appendix~\ref{apx:dataset}. Our experiments are conducted on several open-source instruction-tuned models at varying scales, including Llama 3-Instruct series (1B, 3B, 8B)~\citep{touvron2023llama}, Qwen 2.5-Instruct (1.5B, 3B, 7B)~\citep{qwen2.5}, and Mistral-Instruct (8B)~\citep{mistral2024ministraux}.\footnote{Due to OpenAI API restrictions on logits (limited to the top-20 tokens), we cannot obtain full-vocabulary logits for API-based models. Exploring methods compatible with these restrictions is left for future work.}

\begin{table*}[!t]
\center \small
\resizebox{0.9\linewidth}{!}{%
\begin{tabular}{lllcccccc} \toprule
\textbf{Dataset}                       & \textbf{Model}                                        & \textbf{Method}            & \textbf{Gender} & \textbf{Race} & \textbf{Religion} & \textbf{Disability} & \textbf{Age} & \textbf{Average} \\ \midrule
\multirow{12}{*}{GSM8K}        & \multirow{4}{*}{Llama-8B}     & ZS-CoT & 73.16 & 73.16 & 73.16   & 73.16   & 73.16 & 73.16 \\
          &             &  Persona\textsubscript{average}      & 76.93 & 70.20                    & 72.20   & 64.86   & 67.06                   & 70.25                    \\
          &             &  Persona\textsubscript{majority} & 79.40 & 76.83                    & 80.46   & 64.87   & 67.63                   & 73.84                    \\
          &             & CHOIR              & \textbf{82.11} & \textbf{81.05}                    & \textbf{82.56}   & \textbf{77.56}   & \textbf{78.39}                   & \textbf{80.33}$^{*}_{\text{\textcolor{Green}{+6.49}}}$                   \\  \cmidrule{2-9}
          &              & ZS-CoT        & 36.54 & 36.54                    & 36.54   & 36.54   & 36.54                   & 36.54                    \\
          &            & Persona\textsubscript{average}      &48.14 & 47.46                    & 46.58   & 63.31   & 63.34                   & 53.77                    \\
          && Persona\textsubscript{majority} & 53.75 &51.78                    &54.01   &63.74   &63.05                   &57.27                    \\ 
          &  \multirow{-4}{*}{Qwen-7B}          &CHOIR              &\textbf{61.79} &\textbf{62.70}                    &\textbf{62.09}   &\textbf{67.55}   & \textbf{70.66}                   & \textbf{64.96}$^{*}_{\text{\textcolor{Green}{+7.69}}}$                    \\ \cmidrule{2-9}
          & \multirow{4}{*}{Mistral-8B} & ZS-CoT        & 84.99 & 84.99                    & 84.99   & 84.99   & 84.99                   & 84.99                    \\
          &             &  Persona\textsubscript{average}      & 82.89 & 84.33                    & 84.53   & 83.70   & 82.79                   & 83.65                    \\
          &             &  Persona\textsubscript{majority} & 82.94 & 84.79                    & 85.11   & 83.75   & 83.02                   & 83.92                    \\
          &             & CHOIR              & \textbf{85.37} & \textbf{87.04}                    & \textbf{87.95}   & \textbf{86.05}   & \textbf{86.43}                   & \textbf{86.57}$^{*}_{\text{\textcolor{Green}{+1.58}}}$                    \\ \midrule 
\multirow{12}{*}{MultiArith}   &      & ZS-CoT        & 83.00 & 83.00                    & 83.00   & 83.00   & 83.00                   & 83.00                    \\ 
          &             &  Persona\textsubscript{average}      & 89.89 & 80.67                    & 82.56   & 69.08   & 75.17                   & 79.47                    \\ 
          &             &  Persona\textsubscript{majority} & 92.89 & 88.00                    & 92.50   & 70.89   & 75.61                   & 83.98                    \\ 
          &   \multirow{-4}{*}{Llama-8B}          & CHOIR              & \textbf{94.67} & \textbf{93.00}                    & \textbf{94.83}   & \textbf{91.50}   & \textbf{91.67}                   & \textbf{93.13}$_{\text{\textcolor{Green}{+9.16}}}$                    \\ \cmidrule{2-9}
          & \multirow{4}{*}{Qwen-7B}             & ZS-CoT        & 36.33 & 36.33                    & 36.33   & 36.33   & 36.33                   & 36.33                    \\
          &             &  Persona\textsubscript{average}      & 56.17 & 49.33                    & 45.22   & 67.83   & 50.25                   & 53.76                    \\
          &             &  Persona\textsubscript{majority} & 63.89 & 53.94                    & 54.72   & 67.72   & 50.56                   & 58.17                    \\
          &             & CHOIR              & \textbf{71.00} & \textbf{66.17}                    & \textbf{64.83}   & \textbf{72.67}   & \textbf{60.50}                   & \textbf{67.03}$^{*}_{\text{\textcolor{Green}{+8.87}}}$                    \\ \cmidrule{2-9}
          & \multirow{4}{*}{Mistral-8B} & ZS-CoT        & 98.17 & 98.17                    & 98.17   & 98.17   & 98.17                   & 98.17                    \\
          &             &  Persona\textsubscript{average}      & 97.50 & 97.44                    & 97.83   & 97.75   & 96.33                   & 97.37                    \\
          &             &  Persona\textsubscript{majority} & 97.50 & 97.50                    & 97.83   & 97.67   & 96.28                   & 97.36                    \\
          &             & CHOIR              & \textbf{98.33} & \textbf{98.50}                    & \textbf{98.50}   & \textbf{98.33}   & \textbf{98.50}                   & \textbf{98.43}$^{*}_{\text{\textcolor{Green}{+0.27}}}$                    \\ \midrule
\multirow{12}{*}{SingleEq}     & \multirow{4}{*}{Llama-8B}     & ZS-CoT        & 49.21 & 49.21                    & 49.21   & 49.21   & 49.21                   & 49.21                    \\
          &             &  Persona\textsubscript{average}      & 53.28 & 47.70                    & 49.67   & 40.75   & 49.70                   & 48.22                    \\
          &             &  Persona\textsubscript{majority} & 55.38 & 52.03                    & 55.18   & 40.29   & 49.93                   & 50.56                    \\
          &             & CHOIR              & \textbf{57.87} & \textbf{57.09}                    & \textbf{57.87}   & \textbf{54.13}   & \textbf{57.48}                   & \textbf{56.89}$^{*}_{\text{\textcolor{Green}{+6.33}}}$                    \\ \cmidrule{2-9}
          & \multirow{4}{*}{Qwen-7B}             & ZS-CoT        & 64.17 & 64.17                    & 64.17   & 64.17   & 64.17                   & 64.17                    \\
          &             &  Persona\textsubscript{average}      & 66.01 & 57.74                    & 48.29   & 83.76   & 63.29                   & 63.82                    \\
          &             &  Persona\textsubscript{majority} & 74.28 & 64.04                    & 57.15   & 84.19   & 63.65                   & 68.66                    \\
          &             & CHOIR              & \textbf{82.48} & \textbf{81.10}                    & \textbf{77.95}   & \textbf{87.60}   & \textbf{80.12}                   & \textbf{81.85}$^{*}_{\text{\textcolor{Green}{+13.19}}}$                    \\ \cmidrule{2-9}
          & \multirow{4}{*}{Mistral-8B} & ZS-CoT        & 95.87 & \textbf{95.87}                    & \textbf{95.87}   & \textbf{95.87}   & \underline{95.87}                   & \textbf{95.87}                    \\
          &             &  Persona\textsubscript{average}      & 95.87 & 95.08                    & 95.01   & 95.18   & 95.28                   & 95.28                    \\
          &             &  Persona\textsubscript{majority} & 95.87 & \underline{95.21}                    & 95.41   & 95.21   & 95.80                   & 95.50                    \\
          &             & CHOIR              & \textbf{96.65} & 95.08                    & \underline{95.47}   & \underline{95.67}   & \textbf{96.06}                   & \underline{95.79}$_{\text{\textcolor{Red}{-0.08}}}$                    \\ \midrule
\multirow{12}{*}{CommonsenseQA} & \multirow{4}{*}{Llama-8B}     & ZS-CoT        & 67.81 & 67.81                    & 67.81   & 67.81   & 67.81                   & 67.81                    \\
          &             &  Persona\textsubscript{average}      & 63.50 & 47.12                    & 53.97   & 58.89   & 53.85                   & 55.47                    \\
          &             &  Persona\textsubscript{majority} & 66.69 & 52.12                    & 61.37   & 58.70   & 54.14                   & 58.60                    \\
          &             & CHOIR              & \textbf{72.97} & \textbf{70.84}                    & \textbf{72.56}   & \textbf{71.01}   & \textbf{70.76}                   & \textbf{71.63}$^{*}_{\text{\textcolor{Green}{+3.82}}}$                    \\ \cmidrule{2-9}
          & \multirow{4}{*}{Qwen-7B}             & ZS-CoT        & 79.52 & 79.52                    & 79.52   & 79.52   & 79.52                   & 79.52                    \\
          &             &  Persona\textsubscript{average}      & 72.86 & 61.37                    & 59.30   & 70.64   & 64.33                   & 65.70                    \\
          &             &  Persona\textsubscript{majority} & 75.84 & 65.90                    & 67.43   & 70.84   & 64.65                   & 68.93                    \\
          &             & CHOIR              & \textbf{81.57} & \textbf{78.87}                    & \textbf{80.59}   & \textbf{80.34}   & \textbf{79.69}                   & \textbf{80.21}$_{\text{\textcolor{Green}{+0.69}}}$                   \\ \cmidrule{2-9}
          & \multirow{4}{*}{Mistral-8B} & ZS-CoT        & 69.94 & 69.94                    & 69.94   & 69.94   & 69.94                   & 69.94                    \\
          &             &  Persona\textsubscript{average}      & 68.41 & 69.10                    & 68.82   & 68.14   & 63.60                   & 67.61                    \\
          &             &  Persona\textsubscript{majority} & 68.91 & 71.06                    & 69.78   & 68.06   & 63.80                   & 68.32                    \\
          &             & CHOIR              & \textbf{71.74} & \textbf{72.81}                    & \textbf{73.14}   & \textbf{71.74}   & \textbf{71.42}                   & \textbf{72.17}$^{*}_{\text{\textcolor{Green}{+2.23}}}$         \\
\bottomrule          
\end{tabular}%
}
\caption{\textbf{Performance of CHOIR across five demographic personas (\textit{Gender}, \textit{Race}, \textit{Religion}, \textit{Disability}, and \textit{Age}) on arithmetic and commonsense reasoning datasets.} The ``Average'' column reports mean accuracy (\%) across all demographic personas. Results are averaged over three runs with different random seeds; standard deviations are provided in the Appendix. \textbf{Bold} values indicate the best performance, and $^*$ marks statistically significant improvements ($p<0.05$) based on paired \textit{t}-tests across demographic scores. Subscript values show gains over the strongest baseline. CHOIR enhances accuracy and exhibits harmonized reasoning across variations.}
\label{tab:main}
\vspace{-10pt}
\end{table*}

\smallskip
\noindent \textbf{Baselines and implementation details.}
Our main comparisons (Table \ref{tab:main}) include three representative baselines: (i) Zero-Shot Chain-of-Thought (ZS-CoT)~\citep{zero_shot_cot}, where the model is prompted with ``Let's think step by step'' to elicit reasoning; (ii) Persona\textsubscript{average}, which averages the predictions of persona-assigned LLMs across different demographic variants; and (iii) Persona\textsubscript{majority}, which aggregates predictions by majority vote. We follow the demographic attributes from Table \ref{tab:demographics}. For our main experiments, each primary persona is perturbed along \textit{Gender}, \textit{Race}, and \textit{Religion} (three variants each), and along \textit{Disability} and \textit{Age} (two variants each). 
A robustness analysis on the number of attributes is provided in the Appendix (Figure \ref{fig:fourrace}).
In cases where majority voting produces ties (e.g., three different answers), we randomly select one outcome. To ensure robustness, we report the mean and standard deviation across three runs with different random seeds. We use greedy decoding with a temperature of 0, following prior work~\citep{zero_shot_cot, auto_cot}. For smaller models (below 3B), we use a maximum of 512 tokens to improve reasoning capacity, while larger models (3B and above) use 256 tokens. For our main CHOIR experiments, we set the knowledge weight $\lambda^{(0)}=n-1$ to balance its influence against the $n-1$ counterfactual personas, ensuring that the aggregated demographic variants do not overwhelm the pre-trained reasoning. This ensures a strong and consistent setting for comparison, particularly when model-specific tuning is not feasible. For all individual personas, we set $\lambda^{(i)} = 1$ for $i \ge 1$. We then conduct a model-specific analysis of the $\lambda^{(0)}$ in \S\ref{ssec:rq3}.  Additional baselines, such as Zero-Shot prompting, Few-Shot CoT, Role-Play prompting~\citep{kong-etal-2024-better}, and Self-Consistency~\citep{selfcon} are evaluated on GSM8K with Llama-8B and presented in Appendix~\ref{apx:baselines}.

\section{Experiments}
\label{sec:experiments}


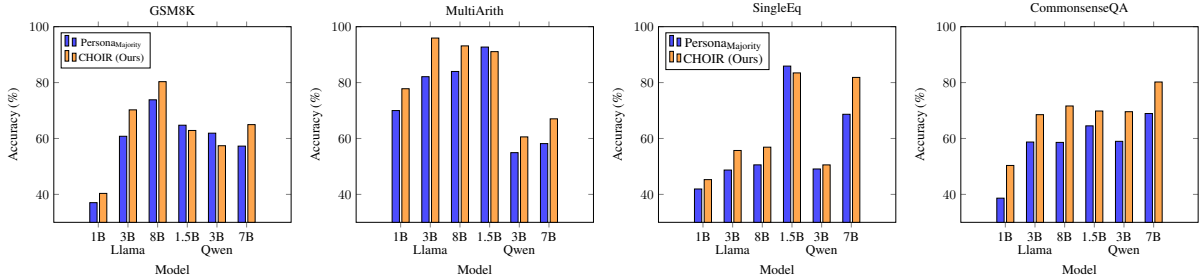
\begin{figure*}[!t]
\centering
\resizebox{\textwidth}{!}{
\begin{tabular}{cccc}

\begin{tikzpicture}
\begin{axis}[
    ybar,
    title={GSM8K},
    ylabel={Accuracy (\%)},
    xlabel={Model},
    symbolic x coords={L1,L3,L8,Q15,Q3,Q7}, 
    xtick=data,
    xticklabels={
        \stackunder{1B}{},
        \stackunder{3B}{Llama},
        \stackunder{8B}{},
        \stackunder{1.5B}{},
        \stackunder{3B}{Qwen},
        \stackunder{7B}{}
    },
    ymin=30, ymax=100,
    bar width=6pt,
    enlarge x limits=0.3,
    legend style={font=\small},
    legend pos=north west,
]

\addplot[fill=blue!70] coordinates {(L1,37.04) (L3,60.79) (L8,73.84) (Q15,64.75) (Q3,61.90) (Q7,57.27)};

\addplot[fill=orange!70] coordinates {(L1,40.35) (L3,70.24) (L8,80.33) (Q15,62.87) (Q3,57.41) (Q7,64.96)};

\addlegendentry{Persona\textsubscript{Majority}}
\addlegendentry{CHOIR (Ours)}
\end{axis}
\end{tikzpicture}
&

\begin{tikzpicture}
\begin{axis}[
    ybar,
    title={MultiArith},
    ylabel={Accuracy (\%)},
    xlabel={Model},
    symbolic x coords={L1,L3,L8,Q15,Q3,Q7},
    xtick=data,
    xticklabels={
        \stackunder{1B}{},
        \stackunder{3B}{Llama},
        \stackunder{8B}{},
        \stackunder{1.5B}{},
        \stackunder{3B}{Qwen},
        \stackunder{7B}{}
    },
    ymin=30, ymax=100,
    bar width=6pt,
    enlarge x limits=0.3,
]

\addplot[fill=blue!70] coordinates {(L1,69.99) (L3,82.11) (L8,83.98) (Q15,92.71) (Q3,54.93) (Q7,58.17)};
\addplot[fill=orange!70] coordinates {(L1,77.80) (L3,95.93) (L8,93.13) (Q15,91.07) (Q3,60.57) (Q7,67.03)};

\end{axis}
\end{tikzpicture}
&

\begin{tikzpicture}
\begin{axis}[
    ybar,
    title={SingleEq},
    ylabel={Accuracy (\%)},
    xlabel={Model},
    symbolic x coords={L1,L3,L8,Q15,Q3,Q7},
    xtick=data,
    xticklabels={
        \stackunder{1B}{},
        \stackunder{3B}{Llama},
        \stackunder{8B}{},
        \stackunder{1.5B}{},
        \stackunder{3B}{Qwen},
        \stackunder{7B}{}
    },
    ymin=30, ymax=100,
    bar width=6pt,
    enlarge x limits=0.3,
    legend pos=north west,
]

\addplot[fill=blue!70] coordinates {(L1,41.93) (L3,48.71) (L8,50.56) (Q15,85.90) (Q3,49.09) (Q7,68.66)};
\addplot[fill=orange!70] coordinates {(L1,45.28) (L3,55.71) (L8,56.89) (Q15,83.46) (Q3,50.55) (Q7,81.85)};

\addlegendentry{Persona\textsubscript{Majority}}
\addlegendentry{CHOIR (Ours)}
\end{axis}
\end{tikzpicture}
&

\begin{tikzpicture}
\begin{axis}[
    ybar,
    title={CommonsenseQA},
    ylabel={Accuracy (\%)},
    xlabel={Model},
    symbolic x coords={L1,L3,L8,Q15,Q3,Q7},
    xtick=data,
    xticklabels={
        \stackunder{1B}{},
        \stackunder{3B}{Llama},
        \stackunder{8B}{},
        \stackunder{1.5B}{},
        \stackunder{3B}{Qwen},
        \stackunder{7B}{}
    },
    ymin=30, ymax=100,
    bar width=6pt,
    enlarge x limits=0.3,
]

\addplot[fill=blue!70] coordinates {(L1,38.65) (L3,58.73) (L8,58.60) (Q15,64.52) (Q3,58.98) (Q7,68.93)};
\addplot[fill=orange!70] coordinates {(L1,50.35) (L3,68.52) (L8,71.63) (Q15,69.83) (Q3,69.58) (Q7,80.21)};

\end{axis}
\end{tikzpicture}

\end{tabular}
}

\caption{\textbf{Performance of CHOIR vs. Persona\textsubscript{Majority} across four reasoning datasets.} Bars represent average accuracy across \textit{five demographic attributes} for Llama and Qwen models.}
\label{fig:scale-new}
\vspace{-15pt}
\end{figure*}

\subsection{RQ1: Does CHOIR improve reasoning over strong baselines?}

The results in Table \ref{tab:main} provide a clear affirmative answer, demonstrating that CHOIR's improvements are evident in three key ways:

\smallskip
\noindent \textbf{A ``Floor-Raising'' Effect: Elevating Weaker Baselines.} CHOIR can generate strong results even when the individual personas are relatively weak. This is most evident in scenarios where the baselines struggle. For instance, on CommonsenseQA with Llama-8B, the average persona performance is a low 55.47\%, which is even lower than ZS-CoT (67.81\%), indicating significant confusion among the individual personas. CHOIR synthesizes these weaker signals into a strong result of 71.63\%, a +16.16 point absolute improvement. This shows that CHOIR's harmonization is not a simple average but an intelligent fusion that can identify and amplify the correct reasoning signal even among conflicting outputs. 

\smallskip
\noindent \textbf{A ``Ceiling-Pushing'' Effect: Optimizing Strong Baselines.} CHOIR is not only for remedial cases; it also excels at optimizing already strong performance. On MultiArith with Llama-8B, where the Persona\textsubscript{majority} baseline is already a high 83.98\%. CHOIR pushes accuracy even further to 93.13\%, achieving a +9.16 point improvement over the strongest baseline. This demonstrates that even when a simple ensemble method like majority voting is effective, CHOIR's dynamic, logit-level weighting can extract additional signal to solve more complex problems.

\smallskip
\noindent \textbf{Enhanced Robustness and Stability.} Beyond accuracy, CHOIR improves reasoning reliability by addressing the instability and failure modes of baselines: \textbf{First, it mitigates performance variance and sensitivity to demographic terms.} The Persona\textsubscript{average} often exhibits high variance across demographic groups (e.g., $\pm$5.3 for disability using Llama3.1-8B-Instruct on GSM8K). As noted by~\cite{gupta2024bias}, LLMs perform worse for certain personas, such as ``disabled''. However, our findings reveal that replacing ``disabled'' with ``able-bodied'' performed even worse, possibly due to its training data imbalances and associative priors developed during pre-training or post-training and the sensitivity of Persona-Assigned LLMs to minor perturbations analyzed in \S\ref{sec:analysis}. From this aspect, we can see that CHOIR enhances performance regardless of the Persona-Assigned LLM's original capability and mitigates the model's sensitivity towards certain demographics, confirming its robustness and effectiveness. \textbf{Second, CHOIR resolves ambiguous outcomes from simpler ensembles.} The Persona\textsubscript{majority} is also prone to failure when reasoning paths are only a few and diverge completely, resulting in a high number of indecisive ``tied votes'' where no consensus can be reached. By contrast, CHOIR's dynamic fusion can produce a definitive answer, enhancing its overall reliability.

\subsection{RQ2: Does CHOIR's Advantage Generalize?}

Our analysis confirms that CHOIR is a broadly generalizable method whose advantages are maintained or amplified as model capability increases. We break down this analysis into two parts: broad generalization across various contexts and a deeper analysis of performance scaling.

\smallskip
\noindent \textbf{Broad Generalization Across Tasks, Models, Demographics, and Prompt Templates:} As shown in Table \ref{tab:main}, CHOIR's effectiveness is not limited to a single domain or model family. It delivers significant average improvements across different tasks and is robust across Llama, Qwen, and Mistral architectures and various demographics (Gender, Race, Religion, Disability, and Age). In addition, CHOIR is robust to the specific wording of the persona instruction. When applying alternative persona instructions, such as ``Adopt the identity of \{persona\}. Answer the questions while staying in strict accordance with the nature of this identity.'' (as shown in Table~\ref{tab:instruction}), CHOIR still improves the Llama-1B model's performance to 38.9\%, outperforming the baselines. This demonstrates that CHOIR is robust to different persona instructions.

\smallskip
\noindent \textbf{Gains Maintained or Amplified with Model Capability:} A detailed analysis of CHOIR's performance relative to the strong Persona\textsubscript{majority} baseline across model families and scales (Figure \ref{fig:scale-new}) reveals key trends in CHOIR's effectiveness: 
\textbf{(1) Consistent Superiority:} CHOIR outperforms the strong Persona\textsubscript{majority} baseline in the vast majority of cases across all four datasets and the 24 specific model configurations tested. This consistent advantage confirms that CHOIR's dynamic, logit-level harmonization is more effective than a simple voting ensemble. 
\textbf{(2) Maintained or Scaled Advantage:} On CommonsenseQA, for example, the lead over the baseline for Llama models grows from 11.7 to 13.0 points (1B to 8B), and for Qwen models, it more than doubles from 5.3 to 11.3 points (1.5B to 7B). This demonstrates that CHOIR is a highly scalable method that effectively leverages the increased reasoning capabilities of larger models.
\textbf{(3) A Nuanced ``Capability Threshold'':} On the GSM8K dataset, while the simpler majority vote is competitive on smaller Qwen models, CHOIR re-establishes a commanding lead of over 7 points (64.96\% vs. 57.27\%) at the 7B scale. This indicates that as models cross a certain capability threshold, CHOIR's fusion mechanism becomes increasingly crucial for unlocking their better performance.

\subsection{RQ3: What are the key components driving CHOIR's success?}
\label{ssec:rq3}

Ablation studies (Table \ref{tab:ablation}) validate CHOIR's design, confirming that its success relies on several key components:

\smallskip
\noindent \textbf{Diverse Persona Paths are Essential:} The primary source of CHOIR's strength is the complementary knowledge unlocked by different personas. When we remove this diversity by replacing the counterfactual paths with a single, repeated reasoning path (CHOIR w/ Same Reasoning Path), accuracy drops from 82.1\% to 73.0\%, confirming that the complementary knowledge from the different personas is important for performance gain.

\begin{table}[t]
\center \small
\resizebox{0.95\columnwidth}{!}{%
\begin{tabular}{lc}
\toprule
\bf CHOIR Components & \bf Accuracy (\%) \\
\midrule
CHOIR (Full Method) & 82.1 \\
\midrule
\textit{Ablations:} & \\
\quad w/ Same Reasoning Path (wo/ Diversity) & 73.0 \\
\quad w/ Random Personas (wo/ Structure) & 80.3 \\
\quad wo/ Pre-trained Knowledge & 79.3 \\
\bottomrule
\end{tabular}
}
\caption{\textbf{Ablation studies for CHOIR with Llama-8B on GSM8K (Gender).} Results show that CHOIR's performance critically depends on: (1) leveraging \textit{diverse reasoning paths}, (2) generating them through \textit{structured counterfactual personas}, and (3) anchoring the ensemble with the model's \textit{pre-trained knowledge}.}
\label{tab:ablation}
\vspace{-5pt}
\end{table}

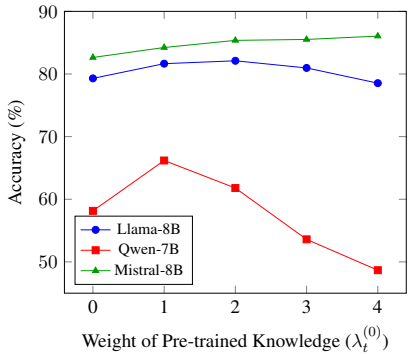
\begin{figure}[!t]
    \center
    \resizebox{0.7\columnwidth}{!}{%
    \begin{tikzpicture}
        \begin{axis}[
            xlabel={Weight of Pre-trained Knowledge ($\lambda^{(0)}_t$)},
            ylabel={Accuracy (\%)},
            xtick={0,1,2,3,4},
            xticklabels={0,1,2,3,4},
            ymin=45, ymax=90,
            ytick={50, 60, 70, 80, 90},
            scaled ticks=false, 
            legend pos=south west,
            legend style={font=\small}
        ]
            \addplot[color=blue, mark=*] coordinates {
                (0, 79.30) (1, 81.65) (2, 82.11) (3, 80.97) (4, 78.54)
            };
            \addlegendentry{Llama-8B}
            
            \addplot[color=red, mark=square*] coordinates {
                (0, 58.15) (1, 66.19) (2, 61.79) (3, 53.60) (4, 48.67)
            };
            \addlegendentry{Qwen-7B}
            
            \addplot[color=green!60!black, mark=triangle*] coordinates {
                (0, 82.64) (1, 84.23) (2, 85.37) (3, 85.52) (4, 86.05)
            };
            \addlegendentry{Mistral-8B}
            
        \end{axis}
    \end{tikzpicture}
    }
    \caption{\textbf{Analysis of the pre-trained knowledge weight $\lambda^{(0)}_t$ on GSM8K (Gender).} Results reveal a model-specific optimal balance, indicating that different architectures have varying reliance on their base priors.}
    \vspace{-15pt}
    \label{fig:weight-new}
\end{figure}

\smallskip
\noindent \textbf{Structured Perturbations are Better than Random Ones:} The way diversity is generated matters. Using our structured, demographically counterfactual personas is more effective than using randomly selected personas from PersonaHub (CHOIR w/ Random Personas) (82.1\% vs. 80.3\%), indicating that CHOIR's controlled, minimal perturbation approach generates more meaningful and complementary reasoning signals.

\smallskip
\noindent \textbf{Balancing Personas with Base Knowledge is Critical:} The model's pre-trained knowledge is vital for grounding the reasoning process. Completely removing the influence of the base model (wo/ Pre-trained Knowledge) causes a significant drop in accuracy from 82.1\% to 79.3\%. While our main experiments use a fixed heuristic $\lambda^{(0)}=n-1$ and achieve significantly improved performance, we conducted a follow-up analysis to investigate the optimal setting for this key hyperparameter across different model architectures by setting the weight range from 0 to 4. As shown in Figure~\ref{fig:weight-new}, we find that the optimal balance is model-specific. While Llama-8B confirms that $\lambda^{(0)}=n-1$ is optimal (82.11\%), Qwen-7B peaks at a weight of 1 (66.19\%), and Mistral-8B's performance is highest at a weight of 4 (86.05\%). This indicates that \textbf{different model families have varying reliance on their base knowledge}. This finding indicates that while CHOIR is highly effective with a default setting, its performance can be further maximized by tuning this hyperparameter.

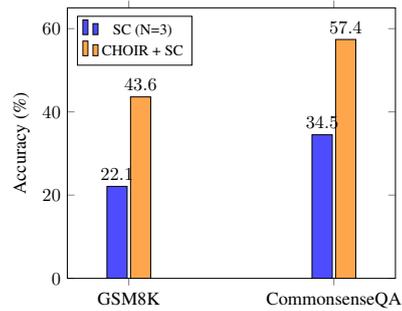
\begin{figure}[!t]
\centering
\resizebox{0.7\columnwidth}{!}{
\begin{tikzpicture}
\begin{axis}[
    ybar,
    bar width=12pt,
    symbolic x coords={GSM8K,CommonsenseQA},
    xtick=data,
    ylabel={Accuracy (\%)},
    ymin=0, ymax=65,
    enlarge x limits=0.3,
    legend style={font=\small},
    legend pos=north west,
    nodes near coords, 
]

\addplot[fill=blue!70] coordinates {(GSM8K,22.1) (CommonsenseQA,34.5)};
\addplot[fill=orange!70] coordinates {(GSM8K,43.6) (CommonsenseQA,57.4)};

\addlegendentry{SC (N=3)}
\addlegendentry{CHOIR + SC}
\end{axis}
\end{tikzpicture}
}
\caption{\textbf{CHOIR + Self-Consistency (SC) vs. SC baseline on GSM8K and CommonsenseQA (Llama-8B, Gender).} Integrating SC reasoning with CHOIR improves accuracy across both datasets.}
\label{fig:sc}
\vspace{-15pt}
\end{figure}

\subsection{RQ4: Can CHOIR enhance other ensemble methods?}

To test the generality of our harmonization mechanism, we applied it to the noisy reasoning paths generated by Self-Consistency (SC). As shown in Figure \ref{fig:sc}, standard SC performs poorly with only three paths, achieving just 22.1\% accuracy. By applying CHOIR's dynamic weighting to these same paths, performance dramatically improves to 43.6\%. This indicates that CHOIR is feasible for integration with existing methods, though it achieves its full potential when the paths are generated from controlled, meaningful perturbations rather than purely stochastic sampling.

\section{Related Work}

\subsection{LLM Reasoning}
LLMs with parameters beyond a certain scale exhibit emergent reasoning abilities \citep{emergent, gsm8k}. When provided with examples of ``chain of thought'' (CoT), which represent intermediate natural language reasoning steps, these models can generate explicit rationales similar to eliciting inductive/abductive reasoning~\citep{cot, reasoning_survey}. Advances such as zero-shot CoT~\citep{zero_shot_cot}, where a model is prompted with phrases like ``Let's think step by step,'' allow for reasoning without the need for explicit few-shot examples. Building on this, multi-step reasoning approaches -- such as Successive Prompting~\citep{successive}, Tree-of-Thought~\citep{tot}, Graph-of-Thought~\citep{got, got2}, Iterating CoT~\citep{iterative_cot}, and Self-Consistency~\citep{self_consistency} -- have become critical strategies for enhancing reasoning in various downstream tasks. Many recent research works have been focused on improving the LLM Reasoning abilities as a pre-training or post-training effort. DeepSeek-R1~\citep{deepseek-r1} is a notable stride forward in the direction of open-source reasoning model trained via large-scale reinforcement learning (RL) without supervised fine-tuning (SFT). There have also been some work in the direction of latent reasoning to scale up the test-time compute via implicitly reasoning in latent space through recurrent depth~\citep{lars, scaling-ttc}.

\subsection{Persona-assigned LLM}
Persona assignment in large language models (LLMs) refers to giving the model a specific role or identity (e.g. “You are a math teacher”) during prompting. This technique, also called \textbf{role prompting} or \textbf{role-play}, has gained popularity as a way to steer LLM behavior in desirable ways. By adopting a persona, an LLM can generate more in-character responses tailored to that role which in turn found to improve the zero-shot reasoning capabilities~\citep{role-play-better-zero-shot}. Coupling a persona with an LLM can activate the model's corresponding parametric knowledge and style, enabling it to produce responses aligned with that role~\citep{two-tales,embodiedplan2025}. Adopting expert personas, especially aligned with the task domain, has demonstrated improved accuracy on mathematical reasoning benchmarks like GSM8K and AQuA~\citep{personamath, roleplay-paradox}. Persona prompts naturally elicit detailed reasoning traces, thereby enhancing inductive, deductive, and multi-step logical reasoning~\citep{role-play-better-zero-shot, lego}. However, inappropriate personas may degrade reasoning performance by introducing extraneous or misleading context. Large-scale studies have shown persona effects to be highly variable, indicating no guaranteed improvements~\cite{personas-not-helpful}. Further, personas might amplify biases and produce unsafe outputs, underscoring the need for cautious implementation~\citep{roleplay-paradox, personas-not-helpful, gupta2024bias,tan-lee-2025-unmasking}. To mitigate biases introduced by personas in LLM reasoning, \citet{doubleedged-sword} propose ensemble prompting, combining persona-driven responses with neutral prompts and selecting the most accurate output through automated evaluation. 

\section{Conclusion}

We reframe persona-induced variance as a source of complementary reasoning rather than bias and introduce CHOIR, a test-time framework that harnesses this diversity to improve performance. Future work includes exploring advanced fusion mechanisms, such as attention-based or learnable weighting, to better integrate reasoning signals.



\section*{Limitations}

The current implementation of CHOIR requires full access to the model's output logits to perform dynamic weighting. This restricts its use to open-source or full-access models. This presents a compelling direction for future research: developing ``logit-free'' harmonization techniques that can approximate CHOIR's dynamic weighting using only top-k probabilities from restricted APIs. This would extend the benefits of collaborative decoding to a wider range of models.


\section*{Ethical Consideration}

Our work operates at the intersection of model reasoning and demographic representation. Our use of demographic terms (e.g., race, gender) is as minimal semantic probes to generate cognitive diversity in the model's reasoning process. A potential risk is that CHOIR, while improving final accuracy, relies on the model's underlying sensitivities to these terms, which may themselves stem from societal biases present in the training data. A primary finding and positive ethical outcome of our work is that CHOIR mitigates performance disparities across different demographic framings. For instance, by improving performance on personas that are otherwise disadvantaged (e.g., the ``disabled'' persona), CHOIR leads to more equitable and reliable model behavior. Instead of amplifying biases, our method harmonizes differences to produce a more robust and stable result that is less sensitive to any single demographic term. As with any technology involving persona assignment, we advocate for its responsible application and further research into its societal impact.

\bibliography{colm2025_conference}

\newpage
\appendix

Code and data are available at \url{https://github.com/dongxiangjue/CHOIR}.

\section{Extended Discussion and Future Directions}

Recent work has shown that persona-driven biases are subtle, persistent, and difficult to remove via simple prompt debiasing~\cite{gupta2024bias}. While much of the existing research treats this performance variation as a flaw to be mitigated, our work takes a fundamentally different approach. CHOIR does not attempt to suppress this variation; instead, it harnesses it, treating the divergent reasoning paths as complementary signals that can be fused into a more robust and accurate final answer. This reframing of sensitivity-as-a-resource is a key contribution of our work.

Our use of minimal counterfactual edits (e.g., pronouns, age, disability terms) is to isolate the effect of demographic variation while preserving semantic consistency. This allowed us to: (1) validate that even minor shifts (e.g., ``he'' vs. ``she'') can alter reasoning; (2) measure persona-induced variation without introducing semantic drift. Beyond counterfactual settings, our ablation results (Table~\ref{tab:ablation}) further confirm that CHOIR’s benefits generalize: even randomly sampled personas (not just counterfactuals) improve performance, surpassing multiple strong baselines. This indicates that CHOIR's robustness is not limited to carefully engineered demographic perturbations but extends to broader forms of reasoning diversity. For future work, extending the framework to richer or more nuanced persona dimensions (e.g., socio-political identity, cultural background, or domain expertise) may reveal additional layers of complementary reasoning, and we leave it to future work. 

For confidence estimation, we employ maximum token probability as a practical proxy due to its simplicity, computational efficiency, and proven stability in prior work. Despite its lightweight nature, this measure consistently yields strong empirical gains. Since CHOIR operates during the final answer aggregation phase of Zero-Shot-CoT, token-level confidence provides a reliable signal for weighting reasoning paths. For future work, richer uncertainty quantification (e.g., sequence-level entropy, calibration-based measures, or cross-persona KL divergence) could further refine CHOIR's weighting mechanism.

\section{More Implementation Details}

This section provides additional details regarding the datasets, models, and prompting strategies used in our study. All experiments are conducted on NVIDIA A5000 24GB GPUs.

\subsection{Dataset and Model Details}
\label{apx:dataset}

The statistics for the datasets are provided in Table \ref{tab:dataset}. The links and licenses for all open-source models are provided in Table~\ref{tab:model-license}.

\begin{table}[!t]
\begin{center}
\resizebox{0.99\linewidth}{!}{
\begin{tabular}{lcccc}
\toprule
\bf Dataset  &\bf Answer Format & \bf $N_{q}$ & \bf $L_{q}$&\bf License \\ \midrule
GSM8K &arabic number &1,319 &46.9  &MIT License \\
SingleEq &arabic number &508 &27.4 &-- \\
MultiArith &arabic number &600 &31.8  &Apache 2.0 \\
CommonsenseQA &option (A-E) &1,221 &27.8  & MIT \\
\bottomrule
\end{tabular}}
\end{center}
\caption{Statistics of datasets. $N_{q}$ denotes the number of questions in each dataset. $L_{q}$ denotes the average words of questions in each dataset.}
\label{tab:dataset}
\end{table}

\begin{table*}[!t]
\centering\small
\resizebox{0.99\linewidth}{!}{
\begin{tabular}{lll} \toprule
\bf Model & \bf Link & \bf License \\
\midrule
\textsc{Llama-3.2-1B-Instruct} & \url{https://huggingface.co/meta-llama/Llama-3.2-1B-Instruct} & Llama 3.2 Community License  \\
\textsc{Llama-3.2-3B-Instruct} & \url{https://huggingface.co/meta-llama/Llama-3.2-3B-Instruct} & Llama 3.2 Community License \\
\textsc{Llama-3.1-8B-Instruct} & \url{https://huggingface.co/meta-llama/Llama-3.1-8B-Instruct} & Llama 3.1 Community License \\
\textsc{Qwen2.5-1.5B-Instruct} & \url{https://huggingface.co/Qwen/Qwen2.5-1.5B-Instruct} & Apache license 2.0 \\
\textsc{Qwen2.5-3B-Instruct} & \url{https://huggingface.co/Qwen/Qwen2.5-3B-Instruct} & qwen-research \\
\textsc{Qwen2.5-7B-Instruct} & \url{https://huggingface.co/Qwen/Qwen2.5-7B-Instruct} & Apache license 2.0 \\
\textsc{Ministral-8B-Instruct-2410} & \url{https://huggingface.co/mistralai/Ministral-8B-Instruct-2410} & mrl \\
\bottomrule
\end{tabular}}
\caption{Model links and licenses.}
\label{tab:model-license}
\end{table*}

\subsection{Prompting Details}

Table \ref{tab:instruction} shows two different templates for assigning personas. For the ``Role-Play'' baseline, we used the specific role-play prompts detailed in Table \ref{tab:role_play}.

\begin{table*}[!t]
\begin{center}
\small
\resizebox{0.98\textwidth}{!}{%
\begin{tabular}{ll}
\toprule
\multicolumn{1}{c}{\bf ID}  &\multicolumn{1}{c}{\bf Instruction}\\ \midrule
1 & You are \{persona\}. Your responses should closely mirror the knowledge and abilities of this persona. \\
2 & Adopt the identity of \{persona\}. Answer the questions while staying in strict accordance with the nature of this identity. \\
\bottomrule
\end{tabular}}
\end{center}
\caption{Two different persona instructions used in our study. To assign a persona to an LLM, we replace \{persona\} in the instruction with the target persona.}\label{tab:instruction}
\end{table*}

\begin{table*}[!t]
\centering
\small
\resizebox{0.99\linewidth}{!}{
\begin{tabular}{p{0.2\textwidth} p{0.8\textwidth}}  
\toprule
\multicolumn{1}{c}{\bf Dataset}  &\multicolumn{1}{c}{\bf Role-play Prompt} \\ 
\midrule
\texttt{aqua, svamp, singleeq, addsub, gsm8k, multiarith} &
\textbf{[role setting:]} From now on, you are an excellent math teacher and always teach your students math problems correctly. And I am one of your students. \newline
\textbf{[reply:]} That's great to hear! As your math teacher, I'll do my best to explain mathematical concepts correctly so that you can understand them easily. \\
\midrule
\texttt{bigbench\_date} &
\textbf{[role setting:]} From now on, you are an excellent teacher and are teaching your students how to calculate dates correctly. I am one of your students. \newline
\textbf{[reply:]} Of course! I'm here to help you with any questions you have about calculating dates correctly. \\
\midrule
\texttt{coin\_flip} &
\textbf{[role setting:]} From now on, you are a coin that always knows which side of your head is facing. \newline
\textbf{[reply:]} Certainly! I'll be your coin for this game. You can go ahead and flip me. \\
\midrule
\texttt{last\_letters} &
\textbf{[role setting:]} From now on, you are an excellent teacher teaching how to create new words by the last letters of several words. \newline
\textbf{[reply:]} Of course! I'd be happy to help you with that task. \\
\midrule
\texttt{object\_tracking} &
\textbf{[role setting:]} From now on, you are a recorder observing how Alice, Bob, and Claire exchange objects. \newline
\textbf{[reply:]} Certainly! I will document the game and tell everyone what they end up with. \\
\midrule
\texttt{commonsenseqa, strategyqa} &
\textbf{[role setting:]} From now on, you are a contestant in a general knowledge quiz who always answers common sense questions accurately. \newline
\textbf{[reply:]} That sounds like a fun challenge! I'm ready to participate and answer your questions. \\
\bottomrule
\end{tabular}
}
\caption{Role-play prompts used in the Role-Play prompting method.}
\label{tab:role_play}
\end{table*}

\subsection{Answer Extraction}

As our experiments are conducted in a zero-shot setting, we adopt the answer extraction methodology from Zero-Shot-CoT~\cite{zero_shot_cot}. Specifically, for each question, we first obtain the answer generated by the LLM. Then, we combine the question, answer, and an answer trigger into a single input for the model. The answer trigger phrases for different response formats are provided in Table \ref{tab:trigger}.

\begin{table}
\centering
\resizebox{\linewidth}{!}{
\begin{tabular}{ll} 
\toprule
\bf Answer Format  & \bf Answer Trigger \\ \midrule

arabic number     & Therefore,
  the answer (arabic numerals) is   \\
option (A-E)    & Therefore,
  among A through E, the answer is  \\
option (A-C)    & Therefore,
  among A through C, the answer is  \\
yes or no       & Therefore,
  the answer (Yes or No) is         \\
string          & Therefore,
  the final answer is               \\
\bottomrule
\end{tabular}}
\caption{Answer triggers for different tasks.}
\label{tab:trigger}
\end{table}

\section{More Results and Analysis}

\subsection{Detailed Performance Analysis by Persona Group}

As shown in Table~\ref{tab:detailed-one}, the quality of the initial persona influences both baseline performance and the degree of improvement under CHOIR. For instance, G(1) starts with the lowest base accuracy ($<$30\%) but sees the largest improvement (+9\%), while the Expert persona starts strong ($\sim$41\%) and improves modestly to 43\%. Despite starting from a lower point, G(1) surpasses G(5) after CHOIR, showing that weaker personas benefit more, while stronger personas still improve. Moreover, a key outcome of the harmonization is that it brings all persona groups into a similar, high-performing range  (39 - 43\%), effectively reducing variance across personas. This illustrates its robustness and its advantage over static persona prompting by mitigating reliance on any single, potentially misaligned persona.

\begin{table*}[!ht]
\centering
\label{tab:more-results}
\resizebox{\textwidth}{!}{%
\begin{tabular}{lcccl} \toprule
\multicolumn{1}{c}{\bf Groups}  &\multicolumn{1}{c}{\bf Gender} &\multicolumn{1}{c}{\bf Accuracy}&\multicolumn{1}{c}{\bf CHOIR}&\multicolumn{1}{c}{\bf Personas}\\ 
\midrule \rowcolor{gray!10}
NA & -- & 37.4 & -- &  \\
\multirow{3}{*}{1}  & he & 29.4 & \multirow{3}{*}{39.1} & a hardworking construction worker tirelessly petitioning for his sibling's immigration case \\
  & she & 29.5 & & a hardworking construction worker tirelessly petitioning for her sibling's immigration case \\
  & they   & 30.8 & & a hardworking construction worker tirelessly petitioning for their sibling's immigration case \\  \rowcolor{gray!10}
  & he & 39.0 &  & a retired basketball coach with decades of experience, known for his expertise in developing point guards \\ \rowcolor{gray!10}
  & she & 40.6 & & a retired basketball coach with decades of experience, known for her expertise in developing point guards   \\ \rowcolor{gray!10}
\multirow{-3}{*}{2}  & they & 40.0 & \multirow{-3}{*}{43.0} & a retired basketball coach with decades of experience, known for their expertise in developing point guards \\
\multirow{3}{*}{3}  & he & 37.6 & \multirow{3}{*}{42.0} & a single man who takes his personal development and emotional health seriously \\
  & she & 37.9 & & a single woman who takes her personal development and emotional health seriously \\
  & they & 37.8 & & a single person who takes their personal development and emotional health seriously \\ \rowcolor{gray!10}
  & he & 39.7 &  & a retired childcare professional who always has a soothing lullaby up his sleeve \\ \rowcolor{gray!10}
  & she & 39.4 & & a retired childcare professional who always has a soothing lullaby up her sleeve \\ \rowcolor{gray!10}
\multirow{-3}{*}{4}  & they & 39.8 & \multirow{-3}{*}{42.2} & a retired childcare professional who always has a soothing lullaby up their sleeve \\
\multirow{3}{*}{5}  & he & 37.8 & \multirow{3}{*}{40.0} & a music producer who collaborates with Bette Midler on his new album \\
  & she & 37.2 & & a music producer who collaborates with Bette Midler on her new album \\
  & they & 37.7 & & a music producer who collaborates with Bette Midler on their new album \\  \rowcolor{gray!10}   
  & he & 41.2 &  & a mathematics professor with his keen interest on arithmetic and reasoning \\ \rowcolor{gray!10}
  & she & 42.5 & & a mathematics professor with her keen interest on arithmetic and reasoning \\ \rowcolor{gray!10}
 \multirow{-3}{*}{Expert}  & they & 41.9 & \multirow{-3}{*}{43.5} & a mathematics professor with their keen interest on arithmetic and reasoning \\ \bottomrule
 
\end{tabular}%
}
\caption{Detailed results for Figure~\ref{fig:one}.}
\label{tab:detailed-one}
\end{table*}

\section{Comparison with Additional Baselines}
\label{apx:baselines}

Table \ref{tab:other} provides a comprehensive comparison of CHOIR against a wider range of baseline methods on the GSM8K dataset using the Llama-8B model. These baselines include standard methods like Zero-Shot and Few-Shot prompting, as well as more advanced reasoning techniques like Role-Play prompting~\citep{kong-etal-2024-better} and Self-Consistency~\citep{selfcon} under Zero-Shot CoT~\citep{zero_shot_cot}. Standard Zero-Shot prompting directly feeds the model a question without additional context, while Few-Shot prompting provides example inputs and outputs to guide responses. Zero-Shot CoT enhances reasoning by appending the phrase ``Let's think step by step,'' encouraging step-by-step reasoning. Role-Play prompting with Zero-Shot CoT is similar to persona-assigned LLMs, prepending carefully crafted role descriptions to generate responses. Self-Consistency generates multiple reasoning paths and then selects the most consistent answer based on majority voting. For Self-Consistency, we set the sampling temperature to $0.7$ and generate three reasoning paths for fair comparisons. 
From the table, we observe that among the baselines, Zero-Shot CoT improves accuracy from 15.5 to 73.2. However, Role-Play, which relies on predefined role prompts, does not generalize well to the math dataset. Notably, standard Self-Consistency with three reasoning paths performs poorly (22.1\%), as the stochastic paths often diverge completely, leading to a high number of tied votes (1,278 tied problems) where no consensus can be reached. This trend aligns with findings from the original Self-Consistency paper~\citep{selfcon}, where using fewer sampled reasoning paths leads to even lower performance than greedy decoding (single-path). As CHOIR consistently surpasses all listed baselines by a substantial margin, we report only the strongest baselines in the main paper for clarity and focus.

\begin{table*}[!t]
\begin{center}
\small
\resizebox{0.8\textwidth}{!}{%
\begin{tabular}{lllllll}
\toprule
& \multicolumn{6}{c}{\bf Accuracy (\%)} \\
\midrule
Zero-Shot & \multicolumn{6}{c}{15.5} \\
Few-Shot & \multicolumn{6}{c}{17.4} \\
Zero-Shot CoT & \multicolumn{6}{c}{73.2} \\
Role-Play & \multicolumn{6}{c}{67.6} \\
Self-Consistency & \multicolumn{6}{c}{22.1} \\  \midrule
&\multicolumn{1}{c}{\textbf{Gender}}&\multicolumn{1}{c}{\textbf{Race}}&\multicolumn{1}{c} {\textbf{Religion}}&\multicolumn{1}{c}{\textbf{Disability}}&\multicolumn{1}{c}{\textbf{Age}}&\multicolumn{1}{c}{\textbf{Average}} \\
\midrule
Persona\textsubscript{average} & 76.9\textsubscript{$\pm$0.5} & 70.2\textsubscript{$\pm$2.2} & 72.2\textsubscript{$\pm$5.0} & 64.9\textsubscript{$\pm$5.3} & 67.1\textsubscript{$\pm$0.4} & 70.3\textsubscript{$\pm$4.7} \\
Persona\textsubscript{majority} & 79.4\textsubscript{$\pm$0.1} & 76.8\textsubscript{$\pm$0.3} & 80.5\textsubscript{$\pm$0.0} & 64.9\textsubscript{$\pm$0.7} & 67.6\textsubscript{$\pm$0.4} & 73.8\textsubscript{$\pm$7.1} \\
CHOIR & \textbf{82.1} & \textbf{81.1} & \textbf{82.6} & \textbf{77.6} & \textbf{78.4} & \textbf{80.3} \\
\bottomrule
\end{tabular}}
\end{center}
\caption{CHOIR performance of Llama3.1-8B-Instruct on GSM8K dataset across five demographic personas.}
\label{tab:other}
\end{table*}

\subsection{Analysis of Computational Overhead}

CHOIR requires generating and processing $N$ persona-conditioned reasoning paths for each input during inference. This comparison remains fair in the main paper, as strong baselines, Persona\textsubscript{average} and Persona\textsubscript{majority}, also aggregate $N$ reasoning paths. In contrast, single-path methods like Zero-Shot, Few-Shot, Zero-Shot-CoT, and Role-Play generate only one reasoning path per query.

In our experiments, the number of counterfactual personas varies by demographic attribute (Table~\ref{tab:demographics}):
$N=3$ for Gender, Race, and Religion, and $N=2$ for Disability and Age.
Table~\ref{tab:trade-off} presents the computational trade-off on GSM8K using LLaMA3-8B-Instruct, comparing CHOIR with both single-path and multi-path baselines in terms of accuracy, latency, and relative efficiency.

CHOIR increases inference latency by approximately $2-3\times$ relative to single-path methods (e.g., Zero-Shot-CoT), as expected from processing multiple reasoning paths. However, it achieves a $1.07-1.12\times$ relative improvement, demonstrating a strong performance-efficiency trade-off.

When compared to multi-path baselines (e.g., Persona\textsubscript{majority}), CHOIR attains substantially higher accuracy than baselines at identical computational cost. For $N=3$, CHOIR is 2.7 points more accurate with no additional latency, underscoring the effectiveness of its dynamic weighting mechanism over static aggregation.

\noindent \textbf{Practical Considerations.} 
In practical deployments, such as medical QA, legal reasoning, or policy analysis, where reasoning reliability outweighs minor latency increases, this trade-off is both justified and beneficial. Future work can explore optimizing CHOIR's efficiency. For instance, adaptive techniques such as ``early exiting'', where computation halts once personas reach a strong consensus, or ``persona pruning'', where only a diverse subset of personas are used for a given query, could reduce latency without sacrificing significant accuracy.

\begin{table*}[!t]
\centering
\small
\resizebox{0.9\textwidth}{!}{%
\begin{tabular}{lccccc} \toprule
\textbf{Method}                          & \textbf{\# Paths} & \textbf{Accuracy (\%)} & \textbf{Latency (s)} & \textbf{Relative Speed} & \textbf{Relative Accuracy} \\ \midrule
Zero-Shot CoT                            & 1                 & 73.2              & 8.2                  & 1.00x               & 1.00x              \\
Persona\textsubscript{majority} (Age)   & 2                 & 67.6              & 16.4                 & 0.50x               & 0.92x              \\
CHOIR (Age)                 & 2                 & 78.4              & 16.4                 & 0.50x               & 1.07x              \\
Persona\textsubscript{majority} (Gender) & 3                 & 79.4              & 24.6                 & 0.33x               & 1.08x              \\
CHOIR (Gender)              & 3                 & 82.1              & 24.6                 & 0.33x               & 1.12x           \\ \bottomrule  
\end{tabular}%
}
\caption{\textbf{Quantitative comparison of accuracy and inference latency on GSM8K (Llama-8B).} The results highlight two key points: (1) CHOIR achieves a significant accuracy boost over single-path methods for a predictable increase in cost, and (2) it is more accurate than the Persona\textsubscript{majority} baseline at the same level of computational overhead.}
\label{tab:trade-off}
\end{table*}

\section{Robustness to the Choice of Demographic Attributes}

\begin{figure*}[!t]
\centering
\resizebox{\textwidth}{!}{
\begin{tabular}{cccc}

\begin{tikzpicture}
\begin{axis}[
    ybar,
    title={GSM8K},
    ylabel={Accuracy (\%)},
    xlabel={Model},
    symbolic x coords={Llama,Qwen,Mistral},
    xtick=data,
    xticklabels={Llama-8B,Qwen-7B,Mistral-8B},
    ymin=50, ymax=100,
    bar width=10pt,
    enlarge x limits=0.3,
    legend style={font=\small},
    legend pos=north west,
]
\addplot[fill=blue!70] coordinates {(Llama,81.05) (Qwen,62.70) (Mistral,87.04)};
\addplot[fill=orange!70] coordinates {(Llama,81.27) (Qwen,61.64) (Mistral,86.81)};
\addlegendentry{Race 1}
\addlegendentry{Race 2}
\end{axis}
\end{tikzpicture}
&

\begin{tikzpicture}
\begin{axis}[
    ybar,
    title={MultiArith},
    ylabel={Accuracy (\%)},
    xlabel={Model},
    symbolic x coords={Llama,Qwen,Mistral},
    xtick=data,
    xticklabels={Llama-3.1-8B,Qwen2.5-7B,Mistral-8B},
    ymin=50, ymax=100,
    bar width=10pt,
    enlarge x limits=0.3,
]
\addplot[fill=blue!70] coordinates {(Llama,93.00) (Qwen,66.17) (Mistral,98.50)};
\addplot[fill=orange!70] coordinates {(Llama,93.17) (Qwen,66.67) (Mistral,98.50)};
\end{axis}
\end{tikzpicture}
&

\begin{tikzpicture}
\begin{axis}[
    ybar,
    title={SingleEq},
    ylabel={Accuracy (\%)},
    xlabel={Model},
    symbolic x coords={Llama,Qwen,Mistral},
    xtick=data,
    xticklabels={Llama-3.1-8B,Qwen2.5-7B,Mistral-8B},
    ymin=50, ymax=100,
    bar width=10pt,
    enlarge x limits=0.3,
]
\addplot[fill=blue!70] coordinates {(Llama,57.09) (Qwen,81.10) (Mistral,95.08)};
\addplot[fill=orange!70] coordinates {(Llama,56.10) (Qwen,81.10) (Mistral,95.28)};
\end{axis}
\end{tikzpicture}
&

\begin{tikzpicture}
\begin{axis}[
    ybar,
    title={CommonsenseQA},
    ylabel={Accuracy (\%)},
    xlabel={Model},
    symbolic x coords={Llama,Qwen,Mistral},
    xtick=data,
    xticklabels={Llama-3.1-8B,Qwen2.5-7B,Mistral-8B},
    ymin=50, ymax=100,
    bar width=10pt,
    enlarge x limits=0.3,
]
\addplot[fill=blue!70] coordinates {(Llama,70.84) (Qwen,78.87) (Mistral,72.81)};
\addplot[fill=orange!70] coordinates {(Llama,71.83) (Qwen,80.18) (Mistral,73.38)};
\end{axis}
\end{tikzpicture}

\end{tabular}
}

\caption{Accuracy comparison on four reasoning datasets. Blue bars represent Race 1 scores and orange bars represent Race 2 scores across three models (Llama-8B, Qwen-7B, Mistral-8B).}
\label{fig:fourrace}
\end{figure*}
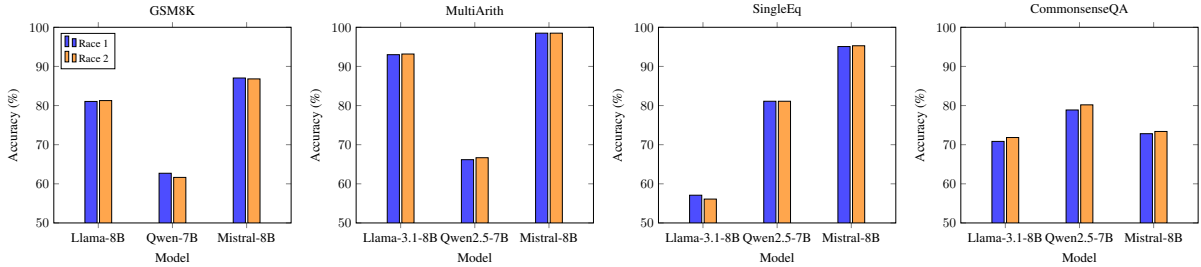

To test CHOIR's robustness to the specific composition of the demographic terms, we compared our primary ``Race 1'' configuration (three attributes: White, Black, Asian) against a ``Race 2'' configuration (four attributes: White, Black, Asian, Hispanic). As shown in Figure~\ref{fig:fourrace}, the performance is highly stable across both sets. This confirms that CHOIR is not overly sensitive to the specific choice or number of demographic attributes used.

\section{Case Study}

Table~\ref{tab:case-study} shows that different demographic personas produce different reasoning paths, leading to different answers, making it difficult to reach a consensus through simple majority voting.

\begin{table*}[!t]
\centering
\small
\resizebox{0.99\textwidth}{!}{%
\begin{tabular}{p{1.5cm}p{12cm}} \toprule
Question &  A hand-painted wallpaper costs \$400 at the market. A DIY will saves 20\% after considering the materials cost. If Ethan made his own hand-painted wallpaper, how much was the total cost?          \\ \rowcolor{gray!10}
Persona 1 & First, we need to calculate the 20\% savings on the \$400 wallpaper. That's 20/100 * \$400 = \$80. So, the cost of the materials would be \$400 - \$80 = \$320. Now, if Ethan made his own hand-painted wallpaper, he would save the \$80 he would have spent on the wallpaper itself. So, the total cost for Ethan would be the materials cost plus the savings, which is \$320 + \$80 = \textcolor{Red}{\$400}. \\
Persona 2 & First, we need to calculate the savings on the wallpaper if Ethan makes it himself. The savings are 20\% of \$400, which is \$400 * 0.20 = \$80. Now, we subtract the savings from the original price to find the cost of the DIY wallpaper. So, \$400 - \$80 = \$320. That's the cost of the materials. But we also need to consider the time and effort Ethan puts in, which is priceless. So, the total cost for Ethan making his own hand-painted wallpaper is \textcolor{Green}{\$320}. \\ \rowcolor{gray!10}
Persona 3 & First, we need to calculate the savings on the wallpaper itself. If the wallpaper costs \$400 and DIY saves 20\%, then the savings would be 20\% of \$400. That's \$400 * 0.20 = \$80. So, the cost of the wallpaper after the savings would be \$400 - \$80 = \$320. Now, we need to consider the cost of materials. If DIY saves 20\% on the wallpaper, it means the materials cost is 80\% of the original price. So, the materials cost would be \$400 * 0.80 = \$320. Therefore, the total cost for Ethan to make his own hand-painted wallpaper would be the cost of the wallpaper after savings plus the cost of materials, which is \$320 + \$320 = \textcolor{Red}{\$640}. \\ \bottomrule  
\end{tabular}%
}
\caption{Case study.}
\label{tab:case-study}
\end{table*}

\end{document}